\PassOptionsToPackage{numbers, compress}{natbib}
\documentclass{article}

\usepackage[table,xcdraw]{xcolor}
\usepackage{makecell}

\usepackage[final]{neurips_2025}

\usepackage[utf8]{inputenc} 
\usepackage[T1]{fontenc}    
\usepackage{hyperref}       
\usepackage{url}            
\usepackage{booktabs}       
\usepackage{amsfonts}       
\usepackage{nicefrac}       
\usepackage{microtype}      

\usepackage{amsmath}
\usepackage{amsthm}

\usepackage{float}
\usepackage{graphicx}
\usepackage{tcolorbox}
\usepackage[normalem]{ulem}
\useunder{\uline}{\ul}{} 
\usepackage{subcaption}

\usepackage{pgfplots}
\pgfplotsset{compat=1.18}

\title{Shaking to Reveal: Perturbation-Based Detection of LLM Hallucinations}

\author{
  Jinyuan Luo\textsuperscript{1} \quad
  Zhen Fang\textsuperscript{1} \quad
  Yixuan Li\textsuperscript{2} \quad
  Seongheon Park\textsuperscript{2} \quad
  Ling Chen\textsuperscript{1}\thanks{Corresponding author} \\
  \textsuperscript{1}Australian Artificial Intelligence Institute, University of Technology Sydney \\
  \textsuperscript{2}Department of Computer Sciences, University of Wisconsin-Madison \\
  \texttt{jinyuan.luo@student.uts.edu.au, \{zhen.fang, ling.chen\}@uts.edu.au} \\
  \texttt{\{sharonli, seongheon\_park\}@cs.wisc.edu}
}

\begin{document}

\maketitle

\begin{abstract}
Hallucination remains a key obstacle to the reliable deployment of large language models (LLMs) in real-world question answering tasks. A widely adopted strategy to detect hallucination, known as self-assessment, relies on the model’s own output confidence to estimate the factual accuracy of its answers. However, this strategy assumes that the model’s output distribution closely reflects the true data distribution, which may not always hold in practice. As bias accumulates through the model’s layers, the final output can diverge from the underlying reasoning process, making output-level confidence an unreliable signal for hallucination detection.  In this work, we propose \textbf{S}ample-\textbf{S}pecific \textbf{P}rompting (\textbf{SSP}), a new framework tha t improves self-assessment by analyzing perturbation sensitivity at intermediate representations. These representations, being less influenced by model bias, offer a more faithful view of the model’s latent reasoning process. Specifically, SSP dynamically generates noise prompts for each input and employs a lightweight encoder to amplify the changes in representations caused by the perturbation. A contrastive distance metric is then used to quantify these differences and separate truthful from hallucinated responses. By leveraging the dynamic behavior of intermediate representations under perturbation, SSP enables more reliable self-assessment. Extensive experiments demonstrate that SSP significantly outperforms prior methods across a range of hallucination detection benchmarks. 


\end{abstract}

\section{Introduction}
In recent years, large language models (LLMs) have demonstrated remarkable capabilities in natural language processing tasks~\cite{achiam2023gpt,grattafiori2024llama}. However, the phenomenon of hallucination in their generated text remains a critical challenge. Hallucination refers to instances where the model produces text that is grammatically and logically coherent but lacks factual accuracy or a verifiable basis~\cite{joshi2017triviaqa,lin2021truthfulqa}. This issue significantly hinders the applicability of LLMs in high-precision domains such as healthcare, law, and science~\cite{ji2023survey,liu2024survey}. Consequently, hallucination detection has emerged as a crucial research problem in ensuring the reliability and trustworthiness of LLMs.

A popular strategy for the detection of hallucinations in LLM is self-assessment~\cite{kadavath2022language, duan2024llms, selfcheckgpt}, which typically estimates the factuality of a response by leveraging the confidence in the output of the model. 
While intuitive and easy to implement, empirical studies have found that their effectiveness can degrade in more complex or realistic scenarios~\cite{du2024haloscope, park2025steer, dasgupta2025hallushift}. One potential reason is a mismatch between the model’s predictive distribution and the true data distribution~\cite{ji2023survey}. As biases accumulate across layers, the final output may drift from the model’s internal reasoning, making output-layer confidence an unreliable signal for self-assessment. 
To overcome this limitation, recent work has begun to shift from probing the output to intermediate representations~\cite{du2024haloscope, azaria2023internal, marks2023geometry, yin2024characterizing, chen2024inside}. While intervening at intermediate representations holds promise, performing self-assessment at this level poses significant challenges. Unlike the output layer, where the predicted distribution naturally provides probabilistic interpretations that reflect the model’s confidence in its predictions, intermediate representations lack such explicit interpretability~\cite{bau2018identifying,rogers2021primer}. Consequently, \textit{how to effectively leverage intermediate-layer information for reliable self-assessment} remains the central challenge that this work aims to address.

\begin{figure}[t]
    \centering
    \includegraphics[width=0.83\textwidth]{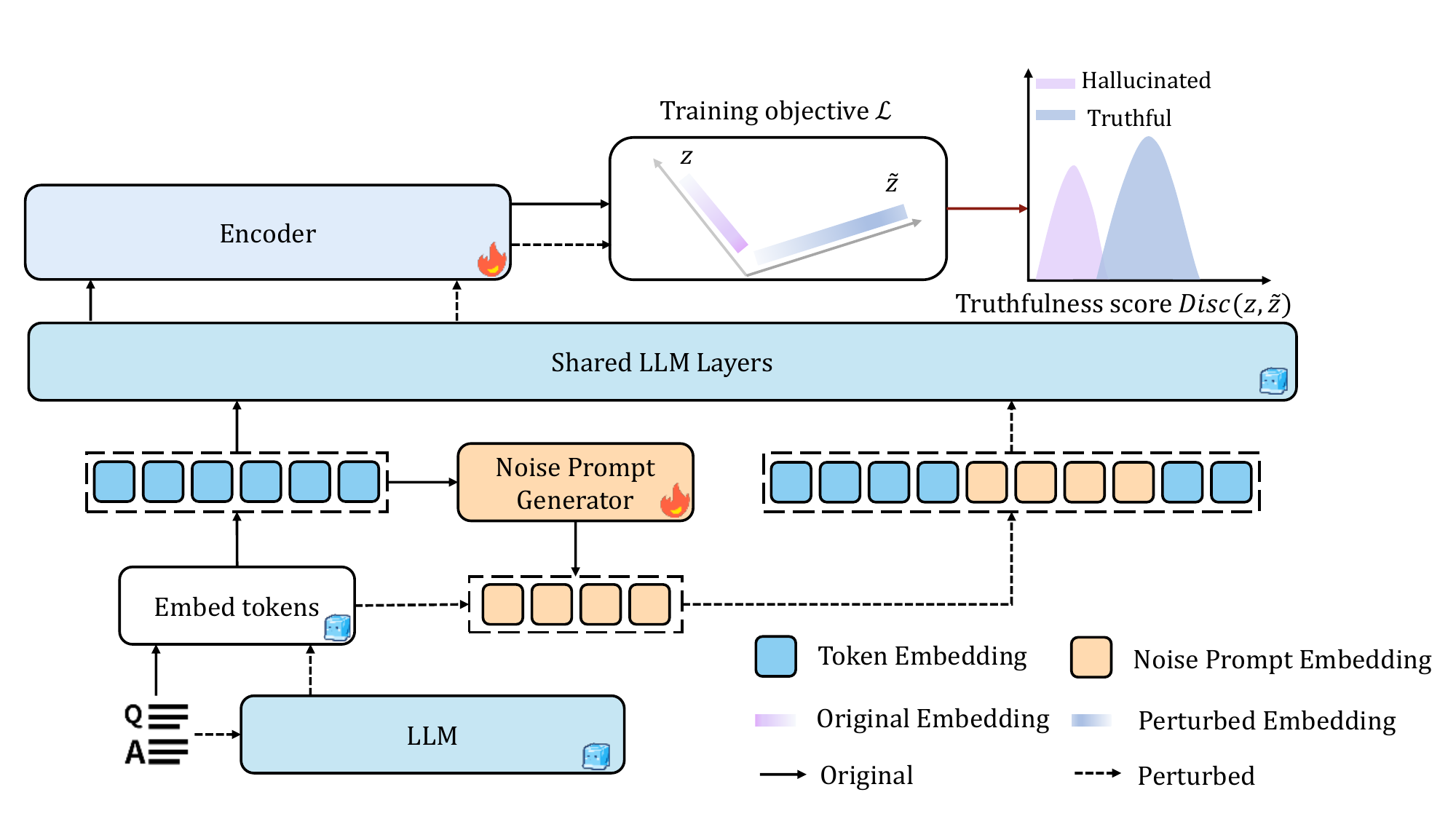}
    \caption{
Overview of \textbf{S}ample-\textbf{S}pecific \textbf{P}rompting (\textbf{SSP}) framework for hallucination detection. Given a question-answer (QA) pair, a noise prompt generator produces a perturbation adapted to the input. The noise prompt is appended to the original answer and passed through a shared LLM backbone to induce representational shifts. The encoder then maps the intermediate representations to a discriminative space and maximize the discrepancy between truthful and hallucinated responses. 
}
    \label{framework}
\end{figure}

In this paper, we propose \textbf{S}ample-\textbf{S}pecific \textbf{P}rompting (\textbf{SSP}), a novel perturbation-based framework that leverages the differential sensitivity of intermediate representations as a signal for hallucination detection. Instead of relying on static or handcrafted prompts, SSP learns to dynamically generate controlled noise prompts tailored to each question–answer pair, inducing perturbations that reveal how internal features respond. 
Our key insight is that truthful and hallucinated responses exhibit distinct representational shifts under input perturbations.
This observation is consistent with ~\cite{meng2022locating,gupta2024rebuilding,zhang2024adversarial,liao2025attack}: factual knowledge is typically encoded in well-structured internal representations, which are tightly coupled with the input and exhibit greater sensitivity in intermediate layers when perturbed, while hallucinated answers remain relatively stable. SSP \emph{amplifies} this signal by introducing a lightweight encoder to extract and compare features before and after perturbation, and by explicitly optimizing a contrastive training objective that encourages larger representation shifts for truthful responses and smaller shifts for hallucinated ones.
In effect, this joint learning of both perturbation prompts and representation encodings enables SSP to be a more effective self-assessment strategy.


Extensive experiments demonstrate the effectiveness  performance of our method across diverse datasets. 
Compared to the state-of-the-art methods, we improve the hallucination detection accuracy by 4.78\% (AUROC) on a challenging TruthfulQA benchmark~\cite{lin2021truthfulqa}. Our results also indicate that SSP generalizes well across different domains. To better understand the role of each component, we conduct comprehensive ablation studies on SSP. The results show that each component contributes to the overall performance.
Our key contributions are summarized as follows:
\begin{itemize}
\item We are the first to leverage the sensitivity of LLMs to input perturbations as a signal for hallucination detection, providing a novel perspective on this problem.

\item We propose \textbf{S}ample-\textbf{S}pecific \textbf{P}rompting (\textbf{SSP}), which generates optimal perturbations for each sample, amplifying the distinction between truthful and hallucinated responses.

\item We conduct ablation studies to evaluate the impact of different components of SSP and demonstrate its effectiveness across diverse LLMs and datasets.
\end{itemize}


\section{Preliminary}
\label{problem setup}

\paragraph{LLM generation probability.}
Let $P_\theta$ denote the conditional probability distribution defined by a pre-trained LLM with parameters $\theta$. Given an input sequence $Q = \{x_1, \dots, x_k\}$ representing the question, where each $x_j$ denotes a token in the input sequence. The model generates an answer $A = \{x_{k+1}, \dots, x_{k+l}\}$ by predicting each token based on the preceding context:
\begin{equation}
P_\theta(x_j \mid x_1, \dots, x_{j-1}), \quad \text{for } j = k+1, \dots, k+l.
\end{equation}
In practice, $A$ is obtained via greedy decoding or beam search to approximate the maximum likelihood output under $P_\theta$~\cite{vaswani2017attention}. The generated answer $A$ is then used, along with the input question $Q$, as the input to a hallucination detector~\cite{du2024haloscope}.

\paragraph{Dataset format.}
Each sample in the dataset consists of a question $Q$, a reference answer $A_{\text{ref}}$, and optionally a context passage (if provided by the dataset). For simplicity, we concatenate the context and the question into a single input sequence, which we denote as $Q$. The dataset can then be represented as:
$\mathcal{S} = \{(Q_1, A_{\text{ref}_1}), \dots, (Q_{n}, A_{\text{ref}_{n}})\}$, where $n$ denotes the total number of samples. 
Given an input $Q$, we use a LLM to generate an answer $A$ in an autoregressive manner~\cite{vaswani2017attention}.
To facilitate hallucination detection, we assign a binary label $y \in \{0,1\}$ to each generated answer $A$, based on its semantic similarity to the reference answer $A_{\text{ref}}$. If $A$ is consistent with $A_{\text{ref}}$, it is labeled as truthful ($y = 1$); otherwise, it is labeled as hallucinated ($y = 0$).
The labeled dataset is defined as
\begin{equation}\label{dataset_format}
    \mathcal{S}_{\text{label}} = \{ (Q_1, A_1, y_1), \dots, (Q_{n}, A_{n}, y_{n})\}.
\end{equation}



\paragraph{Hallucination detection.}
Following the practical setup in recent work~\cite{du2024haloscope}, we denote the true data distribution over truthful input–generation pairs as $ {P}_{\text{true}}$. Given a generated answer $A$ and its corresponding question $Q$, the aim of hallucination detection is to learn a predictor $G$ such that
\begin{equation}
G(Q,A) = 
\begin{cases}
1, & \text{if } A \sim {P}_{\text{true}}(\cdot |Q)\\
0, & \text{otherwise}
\end{cases}.
\end{equation}
A discussion of related works is provided in Appendix~\ref{related works}.

\section{Motivation: Rethinking Self-evaluation for Hallucination detection}\label{Sec::Mot}

\subsection{Self-evaluation and Its Limitation}

\paragraph{Rethinking Self-evaluation \cite{kadavath2022language}.} 
self-assessment \cite{kadavath2022language, duan2024llms,selfcheckgpt} has emerged as a mainstream strategy for hallucination detection in recent research, which leverages the language model’s own outputs or internal signals to evaluate the factual consistency of its responses. Among these, Self-evaluation~\cite{kadavath2022language}, a highly representative method, appends an evaluative prompt $T$, \textit{``Is the proposed answer: (\textbf{A}) True (\textbf{B}) False The proposed answer is''}, to the original question-answer pairs $(Q,A)$ and estimates the confidence of the response by extracting the probability distribution over the subsequent tokens. The probability is then interpreted as the model’s internal belief in the truthfulness of its own answer. 
This method leverages the characteristic that language models tend to produce well-calibrated token probabilities that reflect their internal confidence in a response \cite{guo2017calibration}, formalized as:
\begin{equation*}
P_{\theta}(x=\text{True} \mid Q, A_{\text{Truth}}, T)>P_{\theta}(x=\text{True} \mid Q, A_{\text{Hallu}}, T),
\end{equation*}
which assumes that, given the same question $Q$, the model $P_{\theta}$ assigns higher confidence to factually grounded answers than to hallucinated ones. Ideally, if the model distribution $P_\theta$ perfectly matches the true data distribution ${P}_{\text{true}}$, the principle of Self-evaluation  should remain valid and effective. However, empirical results reported in \cite{du2024haloscope, park2025steer, dasgupta2025hallushift} suggest that its performance may be suboptimal in some specific tasks or scenarios, even under the self-assessment framework. Based on this observation, one possible reason for this limitation \cite{ji2023survey} is the mismatch between the model distribution $P_\theta$ and the true distribution ${P}_{\text{true}}$, which may prevent Self-evaluation from fully achieving its intended effect. 

\paragraph{Towards intermediate-layer Self-evaluation.}
The mismatch between distributions $P_{\theta}$ and ${P}_{\text{true}}$ can arise from various reasons, such as the training objective, model architecture, and algorithm design~\cite{navigli2023biases}. For example, \cite{prato2025large,ren2023self}  argue that a language model's next-token prediction confidence primarily reflects linguistic plausibility rather than factual correctness. Note that the discrepancy between $P_\theta$
and ${P}_{\text{true}}$ often originates not only at the output layer, but also within the model’s intermediate representations.
This issue is further reinforced by the architectural characteristics of mainstream LLMs (e.g., LLaMA-3~\cite{grattafiori2024llama} and Qwen-2.5~\cite{yang2024qwen2}), which are typically composed of multiple stacked Transformer layers.
\cite{schoenholz2016deep,wang2023reducing,glorot2010understanding} demonstrate that as biases propagate through the layers during forward passes, they tend to accumulate and become amplified, ultimately resulting in a significant discrepancy between the model’s predicted distribution  $P_\theta$ and the true distribution ${P}_{\text{true}}$.

Therefore, to effectively mitigate the impact of the bias on Self-evaluation, it is insufficient to focus adjustments solely at the output layer. Instead, Self-evaluation should be designed and optimized at the level of intermediate representations, aiming to suppress the accumulation of bias at its source. This strategy can enhance both the reliability and robustness of the evaluation process. While intervening at intermediate representations holds promise for mitigating the accumulation of distributional bias, performing Self-evaluation at this level poses significant challenges. Unlike the output layer, where the predicted distribution $P_\theta$ naturally provides probabilistic interpretations that reflect the model’s confidence in its predictions, intermediate representations lack such explicit interpretability~\cite{bau2018identifying,rogers2021primer}. Consequently, \textit{how to effectively leverage intermediate-layer information for reliable Self-evaluation} remains the central challenge that this work aims to address.

\subsection{Perturbation-based Self-evaluation}
\label{perturbation}
\paragraph{Perturbation for Self-evaluation.} 
To address the challenge, we begin by making a slight yet essential modification to the standard Self-evaluation to enrich its underlying interpretation. Specifically, inspired by the perturbation strategy~\cite{wallace2019universal,guo2021gradient}, we introduce a noise prompt 
$N$ into the Self-evaluation process and examine the change in the model's confidence before and after the perturbation, i.e.,
\begin{equation}\label{eq::1}
   \Delta P_{\theta}(Q,A,N,T)= \big | P_{\theta}(x=\text{True}|Q,A,T)-P_{\theta}(x=\text{True}|Q,A,N,T) \big |.
\end{equation}

There are two key reasons motivating Eq.~\eqref{eq::1}. \textit{First}, by taking the difference between the predictions before and after perturbation, Eq.~\eqref{eq::1} may partially cancel out certain model-specific biases. As both terms are generated by the same model under similar conditions, shared systematic biases are likely to affect them similarly. This cancellation allows the resulting gap $\Delta P_{\theta}$ to more accurately reflect the change induced by perturbation, rather than being dominated by the model’s inherent bias. \textit{Second}, when $P_{\theta} \approx  {P}_{\text{true}}$, the model’s sensitivity to perturbations can serve as an indicator of its confidence in the predicted answer. For a correct answer $A_{\text{Truth}}$, the model typically exhibits high confidence under standard conditions~\cite{jiang2021can}. Introducing a noise prompt $N$ may disrupt the input context and obscure the key evidence supporting the prediction, leading to a significant drop in the predicted probability:
\begin{equation*}
P_{\theta}(x=\text{True} \mid Q, A_{\text{Truth}}, T) \gg P_{\theta}(x=\text{True} \mid Q, A_{\text{True}}, N, T).
\end{equation*}
In contrast, for a hallucinated answer \( A_{\text{Hallu}} \), where the model generally lacks strong supporting evidence, its confidence remains low or unstable even before perturbation. As a result, adding noise prompt $N$ has limited impact on the prediction. Based on above reasons, we expect the confidence gap induced by perturbation to satisfy the following relationship: 
\begin{equation}\label{Noise}
\text{there exist some noise prompts $N$ such that}~\Delta P_{\theta}(Q, A_{\text{Truth}}, N, T) > \Delta P_{\theta}(Q, A_{\text{Hallu}}, N, T).
\end{equation}

\paragraph{Self-evaluation at internal representations.} 
Beyond its role as a probability gap at the output layer, 
$\Delta P_{\theta}(Q, A, N, T)$
also admits a broader interpretation as a measure of \textit{perturbation-induced change}. 
This interpretation is not restricted to the output probabilities and can be naturally extended to the internal representations of the model. 
In particular, it motivates us to examine how intermediate-layer features respond to input perturbations, 
providing a pathway to generalize Self-evaluation beyond the output layer. Assume that the internal representation is denoted by $E_{\theta}(\cdot)\in \mathbb{R}^d$. We then consider the perturbation-induced representation gap:
\begin{equation}\label{RP}
\Delta E_{\theta}(Q,A,N,T)= \mathbf{Disc}\Big (E_{\theta}(x=\text{True}|Q,A,T),E_{\theta}(x=\text{True}|Q,A,N,T)\Big),
\end{equation}
where $\mathbf{Disc}(\cdot,\cdot)$ is the measure to estimate the difference between the representations $E_{\theta}(x=\text{True}|Q,A,T)$ and $E_{\theta}(x=\text{True}|Q,A,N,T)$. Similar to Eq.~\eqref{Noise}, we expect the representation gap induced by perturbation to satisfy the following relationship:
\begin{equation}\label{Gap}
\text{there exist some noise prompts } N \text{ such that } \Delta E_{\theta}(Q, A_{\text{Truth}}, N, T) > \Delta E_{\theta}(Q, A_{\text{Hallu}}, N, T).
\end{equation}
In the next section, we will introduce how to learn the noise prompt $N$.

\section{Methodology}
Following the motivation in Section~\ref{Sec::Mot}, we introduce  \textbf{S}ample-\textbf{S}pecific \textbf{P}rompting (\textbf{SSP}). 
An overview of the SSP framework is provided in Figure~\ref{framework}.

\subsection{Discrepancy Function}
\label{Discrepancy}
Based on the perturbation framework introduced in Section~\ref{perturbation}, we describe how to extract and compare intermediate-layer representations under input perturbations.
We divide the feature extraction process into two steps. \textit{First}, we extract intermediate-layer representations from the original input $(Q,A,T)$ and perturbed input $(Q,A,N,T)$. \textit{Second}, to amplify the discrepancy between truthful and hallucinated responses under perturbation, we introduce a shared and learnable encoder module \( f_\phi(\cdot) \in \mathbb{R}^d \), where \( \phi \) denotes its trainable parameters, which maps both original and perturbed intermediate representations into the same latent space.
The encoder is designed to preserve the feature information of the LLM embeddings while amplifying the discrepancy  between truthful and hallucinated responses.
As a result, the original and perturbed representations are mapped into:
\begin{equation}\label{representation}
z=f_{\phi}(E_{\theta}(x=\text{True}|Q,A,T))  , \quad \widetilde{z}=f_{\phi}({E}_{\theta}(x=\text{True}|Q,A,N,T)).
\end{equation}
To quantify the magnitude of representation change before and after perturbation, we adopt cosine similarity as the measure. 
 \cite{chen2020simple, zhang2019bertscore, he2020momentum} have demonstrated that cosine-based metrics are robust to variations in feature magnitudes across different layers. Based on this, we define the discrepancy measure as:
\begin{equation}\label{cos_disc}
\mathbf{Disc}(z, \widetilde{z}) = 1 - \cos(z, \widetilde{z} ) = 1 - \frac{z \cdot \widetilde{z}}{|z||\widetilde{z}|}.
\end{equation}
 According to Eq.~\eqref{Gap}, 
the cosine similarity between $z$ and $\widetilde{z}$ is lower for truthful answers, leading to higher discrepancy values compared to hallucinated responses. Formally, we expect that: $\mathbf{Disc}( z_{\text{Truth}} , \widetilde{z}_{\text{Truth}} ) > \mathbf{Disc}( z_{\text{Hallu}}, \widetilde{z}_{\text{Hallu}} )$.
We also compare several alternative distance metrics (e.g., Euclidean distance~\cite{danielsson1980euclidean}, Manhattan distance~\cite{malkauthekar2013analysis}), and find that cosine-based discrepancy achieves the best empirical separability. Detailed results are presented in Table~\ref{ablation::discrepancy}.
\subsection{Sample-Specific Noise Prompt Generation}


\paragraph{Initialization of noise prompts.}
 Considering that each question–answer pair $(Q,A)$ carries sample-specific characteristics, we initialize a unique noise prompt $N$ for each sample. Specifically, given $(Q, A)$, we guide the LLM to dynamically generate a corresponding $N$, formalized as:
\begin{equation} \label{seed_prompt}
    N \sim P_\theta(x \mid \text{SeedPrompt}, Q, A).
\end{equation}
The SeedPrompt is an instruction designed to guide the generation of a natural language sentence that alters the stylistic tone without affecting the contextual semantics.
To maintain factual consistency, we impose a semantic preservation constraint on the generation of $N$, requiring that it does not introduce semantic contradictions (see Appendix~\ref{seedprompt} for details).
The generated noise prompt $N$ is appended to the end of the answer $A$ as its initialization, forming the perturbed input sequence $(Q,A,N,T)$.

\paragraph{Sample-specific prompt learning.}
However, relying solely on LLM-generated noise prompts $N$ may not produce the optimal perturbations. To address this, we introduce a Sample-specific prompt learning strategy that dynamically optimizes the noise prompt $N$ for each sample to maximize the perturbation-induced changes in intermediate representations.
In implementation,  we first extract the sentence embedding $\textbf{h}$ for each input by applying the LLM's token-embedding layer $\text{Emb}_\theta(\cdot)$. Note that $ \text{Emb}_\theta(\cdot)$ is part of the pre-trained LLM and remains frozen during the training process. We have
\begin{equation}
    \textbf{h}=\text{Emb}_\theta{ (Q,A,T) }.
\end{equation}
To dynamically optimize the noise prompt $N$ based on the input pair $(Q,A)$, we introduce a lightweight prompt generator $\text{M}_\varphi(\cdot)$, implemented as a two-layer MLP. Here, $\varphi$ denotes the trainable parameters of the generator. Despite being learnable, $\text{M}_\varphi$ introduces no significant overhead.
Specifically, the embedding of the noise prompt $N$ is updated as:
\begin{equation}
    \text{Emb}_\theta(N) = \text{M}_\varphi(\mathbf{h}) + \text{Emb}_\theta(N).
\end{equation}
After updating \(\text{Emb}_\theta(N)\), we concatenate it with the original input embeddings. Formally, the new embedding sequence is constructed as:
\begin{equation}
    \text{Emb}_\theta(Q, A, N, T) = \text{Emb}_\theta(Q, A) \oplus \text{Emb}_\theta(N) \oplus \text{Emb}_\theta(T),
\end{equation}
where $\oplus$ denotes the concatenation operation along the sequence dimension.
We then feed the combined sequence back into the LLM for forward propagation. Following the procedure described in Section~\ref{Discrepancy}, we extract both the original representation $z$ and the perturbed representation $\widetilde{z}$  from the intermediate layers for training and hallucination detection.


\subsection{Training Objective}
Based on the definition of the discrepancy function in Eq.~\eqref{cos_disc}, we design a contrastive training loss that encourages larger perturbation-induced representation changes for truthful responses while maintaining smaller changes for hallucinated ones. The learnable components, including the MLP $\text{M}_\varphi(\cdot)$ and the encoder  $f_\phi(\cdot)$ , are optimized accordingly through this objective.

For samples labeled as truthful $(y=1)$, we aim to amplify the difference between the original and perturbed features. The corresponding loss is defined as:
\begin{equation}
    {\ell}^{(i)}_{\text{Truth}} =  \max(0, \cos(z_i, \widetilde{z}_i) - \tau_{T}),
\end{equation}

where $\tau_T$ is the upper bound threshold on the cosine similarity for truthful responses. 
For samples labeled as hallucinated $(y=0)$, we aim to maintain a high similarity between the original and perturbed features. The corresponding loss is defined as:
\begin{equation}
    {\ell}^{(i)}_{\text{Hallu}} = \max(0, \tau_H - \cos(z_i, \widetilde{z}_i)),
\end{equation}
where $\tau_{H}$ is the lower bound threshold on the cosine similarity for hallucinated responses. 
Both $\tau_T$ and $\tau_H$ are treated as hyper-parameters.
Given the labeled dataset $\mathcal{S}_{\text{label}}$ introduced in Eq.~\eqref{dataset_format}, the final optimization problem can be written as:
\begin{equation}\label{loss}
  \min_{\varphi,\phi}  \frac{1}{n} \sum_{i=1}^{n} \left( y_i \cdot  {\ell}_{\text{Truth}}^{(i)} + (1-y_i) \cdot  {\ell}_{\text{Hallu}}^{(i)} \right).
\end{equation}
\paragraph{Scoring strategy.} 
After training, we use the discrepancy function in Eq.~\eqref{cos_disc} as the scoring mechanism for hallucination detection.
A high discrepancy score indicates that the model’s internal semantics are significantly disturbed by the noise prompt $N$, suggesting more likely truthful response. Based on the scoring  function, the hallucination detector is 
\begin{equation}
G_\lambda(z,\widetilde{z}) = 
\begin{cases}
1, &  \mathbf{Disc}(z,\widetilde{z}) \geq \lambda\\
0, & \text{otherwise}
\end{cases},
\end{equation}
where  $\lambda$ is the threshold for detection.


\section{Experiments}
\label{Experiments}

\subsection{Experimental Setup}\label{setup}
\textbf{Datasets and models.}
We conduct experiments on four generative QA tasks: two open-book QA datasets  CoQA~\cite{reddy2019coqa} and TruthfulQA~\cite{lin2021truthfulqa}; a closed-book QA dataset TriviaQA~\cite{joshi2017triviaqa}; and a reading comprehension dataset TydiQA-GP (English)~\cite{clark2020tydi}. 
Following \cite{du2024haloscope}, we use only 100 labeled QA pairs for training, while keeping the size of the testing set consistent. 
More datasets and implementation details are provided in Appendix~\ref{details}.
We evaluate our method on two families of widely used open-source LLMs that provide accessible internal representations: LLaMA-3-8B-Instruct~\cite{grattafiori2024llama} and Qwen-2.5-7B-Instruct~\cite{yang2024qwen2}. 
By default, text generations are produced using greedy sampling, which selects the most probable token at each decoding step.

\textbf{Baselines.} We evaluate SSP against a diverse set of 12 baseline methods, including existing state-of-the-art. The baselines are categorized as follows: (1) logit-based methods-Perplexity~\cite{ren2022out} and Semantic Entropy~\cite{kuhn2023semantic}; (2) consistency-based methods-Lexical Similarity~\cite{lin2023generating}, SelfCKGPT~\cite{selfcheckgpt} and  EigenScore~\cite{chen2024inside}; (3) prompting-based methods-Verbalize~\cite{lin2022teaching} and Self-evaluation~\cite{kadavath2022language}; and (4) internal state-based methods-Contrast-Consistent Search (CCS)~\cite{burns2022discovering}, HaloScope~\cite{du2024haloscope}, Linear probe~\cite{azaria2023internal}, EarlyDetec~\cite{snyder2024early}, EGH~\cite{hu2024embedding}. To ensure a fair comparison, we assess all baselines on identical test data, employing the default configurations as outlined in their respective papers.

\textbf{Evaluation.} Following previous works~\cite{park2025steer, du2024haloscope}, we evaluate the performance with the area under the curve of the receiver operator characteristic (AUROC). We consider the generation truthful when the similarity score between the generation and the reference answer is larger than a threshold of 0.5. We employ DeepSeek-V3~\cite{liu2024deepseek}, a powerful open-source language model, to compute the similarity between generated answers and reference answers, which is then used to assign evaluation labels as detailed in Appendix~\ref{deepseekv3}. Additionally, following \cite{du2024haloscope}, we show that the results are robust under two alternative similarity metrics—ROUGE~\cite{lin2004rouge} and BLEURT~\cite{sellam2020bleurt}—as detailed in Appendix~\ref{metric}.

\paragraph{Implementation details.}
Following \cite{du2024haloscope,kuhn2023semantic}, we use beam search with 5 beams to generate the most likely answer for evaluation. For baselines that require multiple generations, we sample 10 responses per question using multinomial sampling with a temperature of 0.5. Consistent with~\cite{azaria2023internal,chen2024inside}, we prepend the question to the generated answer and use the embedding of the final token to detect hallucinations. 
We implement the encoder \( f_\phi(\cdot) \) as a three-layer MLP with ReLU activations. 
Then we train the learnable parameters for 40 epochs using the SGD optimizer with an initial learning rate of 0.01. The thresholds $\tau_T$ and $\tau_H$ are set to 0.3 and 0.7, respectively.

\subsection{Main Results} \label{main_results}
As shown in Table~\ref{tab::main_results}, we compare  SSP with competitive hallucination detection methods from the literature. SSP achieves the highest average AUROC score, significantly outperforming other methods on both the LLaMA-3-8B-Instruct and Qwen-2.5-7B-Instruct.
 We observe that SSP outperforms logit-based baselines, exhibiting 11.3\% and 7.78\% improvement over Perplexity and Semantic Entropy on the challenging TruthfulQA task.
From a computational perspective, both logit-based and consistency-based methods incur significant overhead during inference, as they require sampling multiple responses for each question. Following the setting in~\cite{du2024haloscope}, 10 generations per question are used, which leads to substantial computational cost, especially when applied to large-scale datasets. In contrast, SSP only requires computing the representation shift before and after perturbation, making it significantly more efficient during inference.
For prompting-based baselines, accumulated biases in intermediate layers can lead to unreliable confidence estimates, which limits their effectiveness in certain hallucination detection scenarios~\cite{zhou2023navigating}.
Lastly, we compare SSP with internal state-based methods, including CCS, HaloScope, Linear probe, EarlyDetec, and EGH. SSP consistently outperforms all baselines across datasets, achieving the highest average AUROC scores. This demonstrates that our method provides a more reliable signal for hallucination detection.
\begin{figure}[t]
  \centering
  \begin{subfigure}{0.33\textwidth}
    \includegraphics[width=\linewidth]{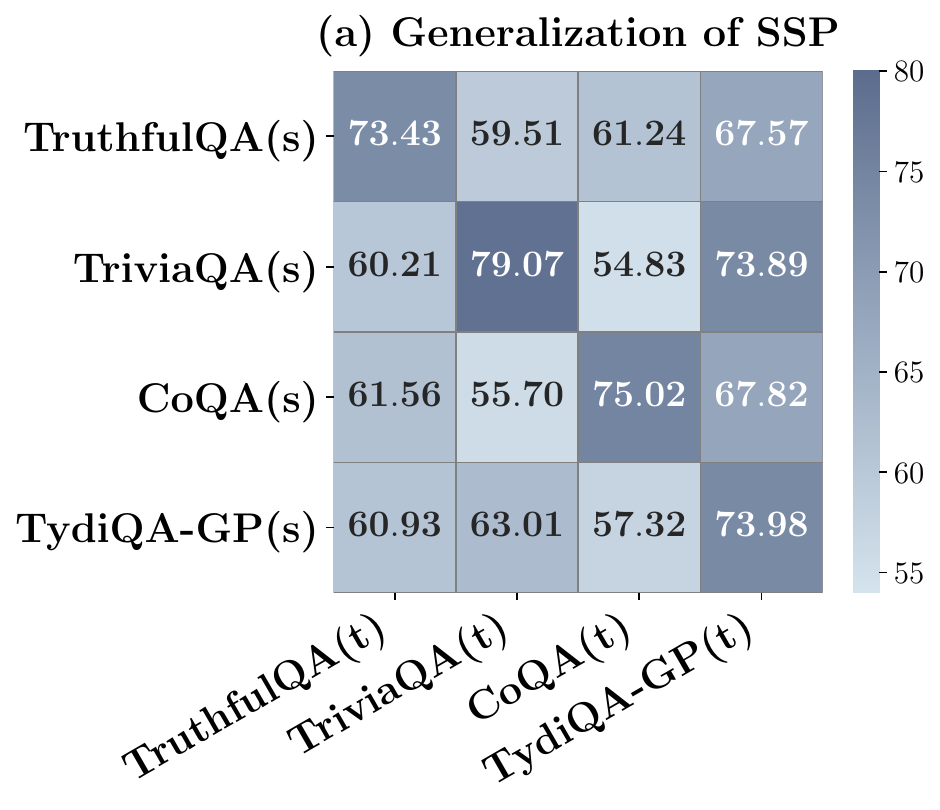}
    \label{fig:ssp}
  \end{subfigure}
  \hspace{-1.5mm}
  \begin{subfigure}{0.33\textwidth}
    \includegraphics[width=\linewidth]{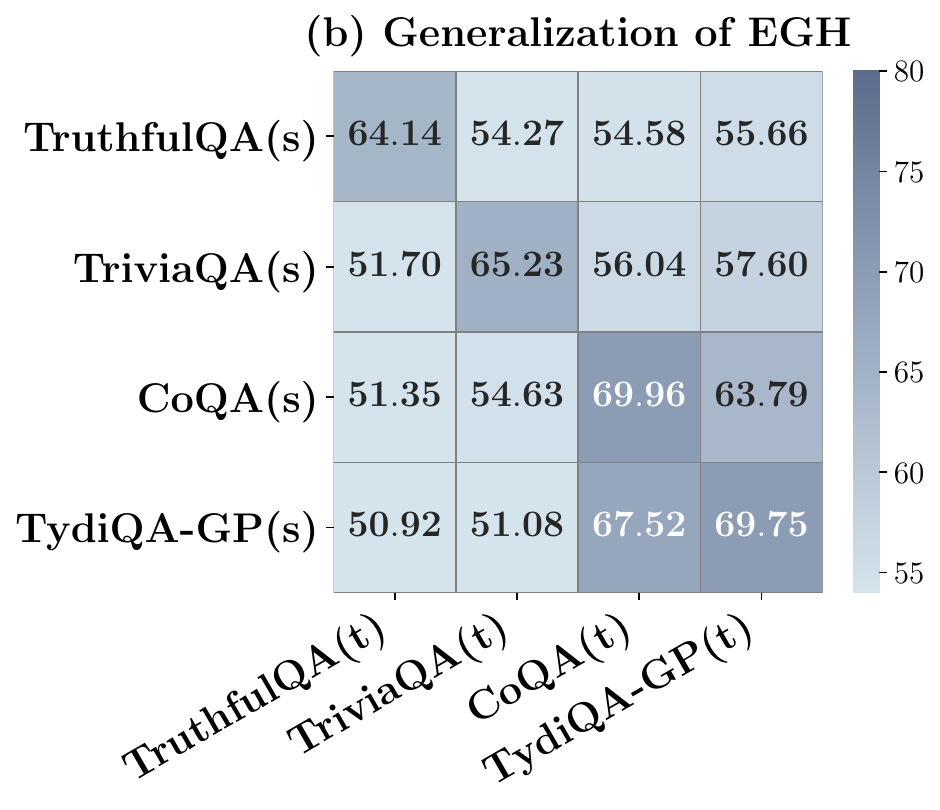}
    \label{fig:egh}
  \end{subfigure}
  \hspace{-1.5mm}
  \begin{subfigure}{0.33\textwidth}
    \includegraphics[width=\linewidth]{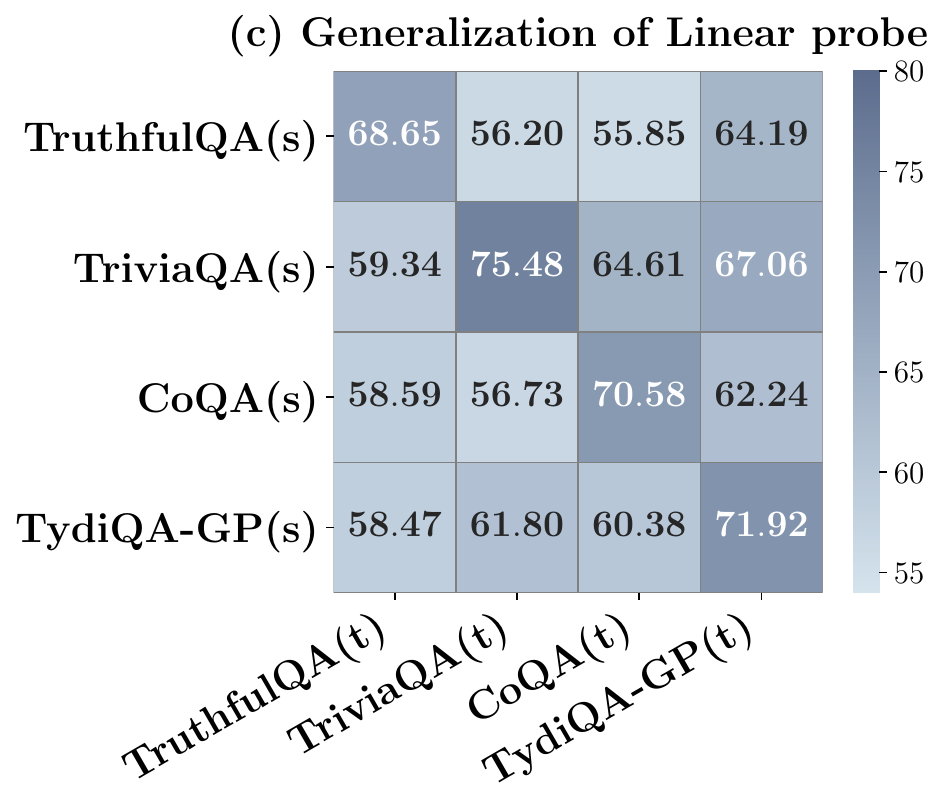}
    \label{fig:linear}
  \end{subfigure}
  \vspace{-0.3cm}
  \caption{Generalization performance comparison across SSP (Ours), EGH~\cite{hu2024embedding}, and Linear probe~\cite{azaria2023internal}. All values are AUROC scores (\%).}
  \label{fig:generalization}
\end{figure}

\begin{table}[t]
\centering
\caption{\textbf{Main results.} Comparison with competitive hallucination detection methods on different datasets. All values are percentages (AUROC, \%). \textbf{Bold} numbers indicate the best performance, and \underline{underlined} numbers indicate the second best.}
{\small
\resizebox{0.8\textwidth}{!}{

\begin{tabular}{c|ccccc|c}
\hline
Model                                         & Method                                      & TruthfulQA                             & TriviaQA                               & CoQA                                   & TydiQA-GP                                            & \textbf{Average}                       \\ \hline
                                              & Perplexity                                  & 62.13                                  & 76.64                                  & 64.87                                  & {\color[HTML]{333333} 53.40}                         & 64.26                                  \\
                                              & Semantic Entropy                            & 58.88                                  & {\ul 78.53}                            & 55.15                                  & {\color[HTML]{333333} 55.21}                         & 61.94                                  \\
                                              & Lexical Similarity                          & 53.64                                  & 78.22                                  & {\ul 77.47}                            & {\color[HTML]{333333} 60.94}                         & 67.57                                  \\
                                              & EigenScore                                  & 56.31                                  & 70.82                                  & 74.30                                  & {\color[HTML]{333333} 72.57}                         & 68.50                                  \\
\multicolumn{1}{l|}{\cellcolor[HTML]{FFFFFF}} & \cellcolor[HTML]{FFFFFF}SelfCKGPT           & \cellcolor[HTML]{FFFFFF}58.74          & \cellcolor[HTML]{FFFFFF}77.56          & \cellcolor[HTML]{FFFFFF}\textbf{78.67} & \cellcolor[HTML]{FFFFFF}{\color[HTML]{333333} 51.29} & 66.57                                  \\
LLaMA-3-8B-Instruct                           & Verbalize                                   & 59.70                                  & 55.43                                  & 53.39                                  & 53.39                                                & 55.48                                  \\
{\color[HTML]{FF0000} }                       & Self-evaluation                             & 53.18                                  & 77.06                                  & 62.30                                  & \textbf{76.69}                                       & 67.31                                  \\
                                              & CCS                                         & 53.91                                  & 58.58                                  & 52.40                                  & {\ul 74.11}                                          & 59.75                                  \\
                                              & HaloScope                                   & 68.40                                  & 63.70                                  & 64.10                                  & 71.10                                                & 66.83                                  \\
                                              & Linear probe                                & {\ul 68.65}                            & 75.48                                  & 70.58                                  & 71.92                                                & {\ul 71.66}                            \\
                                              & EarlyDetec                                  & 67.68                                  & 68.39                                  & 68.23                                  & 70.72                                                & 68.76                                  \\
                                              & EGH                                         & 64.14                                  & 65.23                                  & 69.96                                  & 69.75                                                & 67.27                                  \\
                                              & \cellcolor[HTML]{CDCDCD}\textbf{SSP (Ours)} & \cellcolor[HTML]{CDCDCD}\textbf{73.43} & \cellcolor[HTML]{CDCDCD}\textbf{79.07} & \cellcolor[HTML]{CDCDCD}75.02          & \cellcolor[HTML]{CDCDCD}73.98                        & \cellcolor[HTML]{CDCDCD}\textbf{75.38} \\ \hline
                                              & Perplexity                                  & 53.60                                  & 52.72                                  & 62.03                                  & 51.97                                                & 55.08                                  \\
                                              & Semantic Entropy                            & 64.25                                  & 71.27                                  & 52.35                                  & 50.17                                                & 59.51                                  \\
                                              & Lexical Similarity                          & 57.50                                  & 65.55                                  & 71.62                                  & 61.75                                                & 64.11                                  \\
                                              & EigenScore                                  & 52.67                                  & 68.36                                  & 72.33                                  & 60.97                                                & 63.58                                  \\
                                              & SelfCKGPT                                   & 65.88                                  & 72.36                                  & \textbf{74.18}                         & 56.50                                                & 67.23                                  \\
Qwen2.5-7B-Instruct                           & Verbalize                                   & 54.25                                  & 51.53                                  & 51.86                                  & 52.25                                                & 52.47                                  \\
                                              & Self-evaluation                             & 51.21                                  & 58.97                                  & 52.13                                  & 55.61                                                & 54.48                                  \\
{\color[HTML]{FF0000} }                       & CCS                                         & 53.58                                  & 50.42                                  & 50.32                                  & 54.58                                                & 52.23                                  \\
                                              & HaloScope                                   & 68.10                                  & 63.00                                  & 63.90                                  & 67.00                                                & 65.50                                  \\
                                              & Linear probe                                & {\ul 70.58}                            & 63.15                                  & 68.46                                  & {\ul 69.72}                                          & 67.98                                  \\
                                              & EarlyDetec                                  & 66.99                                  & {\ul 73.13}                            & 67.24                                  & 69.16                                                & {\ul 69.13}                            \\
                                              & EGH                                         & 63.21                                  & 67.96                                  & 70.91                                  & 65.31                                                & 66.85                                  \\
                                              & \cellcolor[HTML]{CDCDCD}\textbf{SSP (Ours)} & \cellcolor[HTML]{CDCDCD}\textbf{72.03} & \cellcolor[HTML]{CDCDCD}\textbf{74.01} & \cellcolor[HTML]{CDCDCD}{\ul 72.43}    & \cellcolor[HTML]{CDCDCD}\textbf{72.40}               & \cellcolor[HTML]{CDCDCD}\textbf{72.72} \\ \hline
\end{tabular}
}
}
\label{tab::main_results}
\end{table}

\subsection{Generalization of SSP}
We evaluate the generalization capability of SSP across datasets with different distributions. Specifically, we directly transfer the learned sample-specific prompt and encoder from a source dataset ``(s)'' and apply them to a target dataset ``(t)'' to compute scores without additional training.
Figure~\ref{fig:generalization} (a) illustrates the strong cross-dataset transferability of our proposed SSP framework. When transferring parameters from TriviaQA to TydiQA-GP, SSP achieves an AUROC of 73.89\% for hallucination detection, which is competitive with the in-domain performance on TruthfulQA (78.64\%).
Figure~\ref{fig:generalization} (b) and (c) show the generalization results of EGH and the linear probe. Both methods exhibit weaker cross-dataset transferability compared to SSP, with notably lower AUROC scores in most off-diagonal entries. For instance, transferring from TriviaQA to TydiQA-GP yields 57.60\% for EGH and 67.06\% for the linear probe, both falling short of SSP's 73.89\% under the same setting. These results indicate that EGH suffers from limited representation generalization, while the linear probe, despite achieving competitive results in some cases, exhibits unstable performance across datasets.

\begin{table}[t]
\centering
\caption{
\textbf{Prompting strategies and component ablations.} AUROC (\%) results on four datasets.
}
\resizebox{0.7\textwidth}{!}{


\begin{tabular}{c|cccc|c}
\hline
Method             & TruthfulQA     & TriviaQA       & CoQA           & TydiQA-GP      & \textbf{Average}                      \\ \hline
Static prompt      & 68.81          & 75.49          & 66.75          & 72.67          & {\color[HTML]{333333} 70.93}          \\
Prompt Tuning      & 70.21          & 76.21          & 66.88          & 73.05          & {\color[HTML]{333333} 71.59}          \\ \hline
SSP w/o Encoder    & 65.87          & 67.03          & 57.90          & 72.47          & {\color[HTML]{333333} 65.82}          \\
SSP w/o SeedPrompt & 72.20          & \textbf{79.95} & 74.21          & 73.44          & {\color[HTML]{333333} 74.95}          \\
\rowcolor[HTML]{CDCDCD} 
SSP                & \textbf{73.43} & 79.07          & \textbf{75.02} & \textbf{73.98} & {\color[HTML]{333333} \textbf{75.38}} \\ \hline
\end{tabular}
}
\label{ablation::component}
\end{table}

\subsection{Ablation Study} \label{ablation}
We conduct detailed ablation studies to investigate the contribution of each component in SSP. Additional ablation results are presented in Appendix~\ref{metric}--\ref{time}. 

\paragraph{Comparison of prompting strategies and SSP components.}\label{component}
We compare five variants to evaluate the impact of prompt design and components on hallucination detection. All experiments are conducted using the LLaMA-3-8B-Instruct model. As shown in Table~\ref{ablation::component}, static prompt achieves a baseline performance of 70.93\% average AUROC across datasets.
Prompt Tuning offers a slight improvement (71.59\%), indicating that global learned prompts can help but are still limited. Removing the encoder from SSP leads to a significant performance drop (65.82\%), confirming its essential role in amplifying representational discrepancy. When the SeedPrompt is removed, performance decreases moderately (74.95\%), showing that the SeedPrompt provides a useful inductive bias. 

\begin{figure}[t]
  \centering
  \begin{subfigure}[t]{0.48\textwidth}
    \centering
    \includegraphics[height=4.5cm]{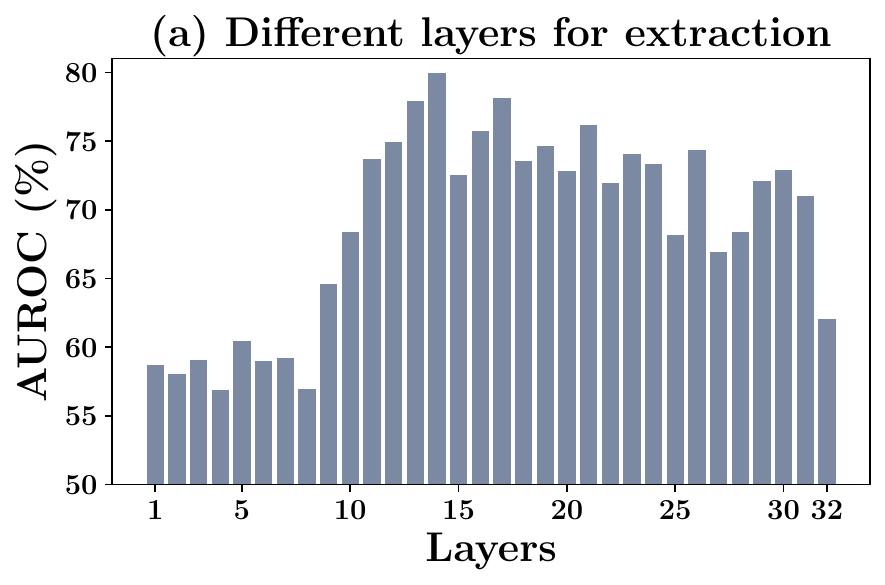}
    \label{fig:triviaqa_layers}
  \end{subfigure}
  \hfill
  \begin{subfigure}[t]{0.48\textwidth}
    \centering
    \includegraphics[height=4.5cm]{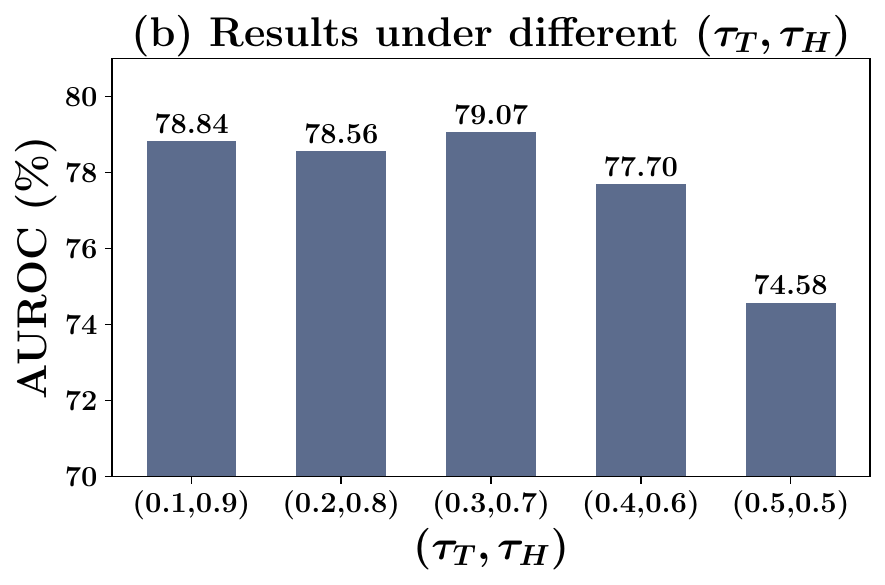}
    \label{fig:threshold_ablation}
  \end{subfigure}
  \caption{Visualization of performance under different layers (left) and threshold settings (right).}
  \label{fig:layer_threshold_combined}
\end{figure}

\paragraph{Impact of layer selection on SSP performance.}
Figure~\ref{fig:layer_threshold_combined} (a) shows hallucination detection results using representations extracted from different layers of the LLM. AUROC scores for classifying truthful and hallucinated responses are computed using the LLaMA-3-8B-Instruct model. All other configurations follow the main experimental setup.  We observe that performance increases with depth up to the middle layers, after which it starts to decline. This trend suggests that the LLM captures meaningful contextual semantics in its middle layers~\cite{azaria2023internal,chen2024inside}.
As representations propagate deeper, accumulated deviations may degrade hallucination detection performance. These results highlight the effectiveness of internal representations in capturing meaningful signals for hallucination detection.

\paragraph{Effect of discrepancy function design.}
We investigate how the design of the discrepancy function influences hallucination detection performance. Specifically, we compare the cosine-based formulation defined in Eq.\eqref{cos_disc} against alternative distance measures, including Manhattan distance~\cite{malkauthekar2013analysis}, Euclidean distance~\cite{danielsson1980euclidean}, and Kullback–Leibler (KL) divergence \cite{kl-divergence}. For each discrepancy function, we define a corresponding score function that computes the magnitude of representation change between the original and perturbed inputs. As shown in Table~\ref{ablation::discrepancy}, the cosine-based metric consistently provides better separability between truthful and hallucinated responses across all evaluated datasets. All experiments in this ablation study are conducted using the LLaMA-3-8B-Instruct model.
\begin{table}[t]
\centering
\caption{
Hallucination detection performance using different discrepancy functions as score metrics. All values are AUROC scores (\%).
}
\resizebox{0.7\textwidth}{!}{

\begin{tabular}{c|cccc|c}
\hline
Method                                                         & TruthfulQA     & TriviaQA       & CoQA           & TydiQA-GP      & \textbf{Average} \\ \hline
Manhattan distance                                                         & 59.18          & 54.21          & 59.31          & 56.99          & 57.42            \\
Euclidean distance                                                        & {\ul 63.60}    & {\ul 72.38}    & {\ul 60.11}    & 59.23          & {\ul 63.83}      \\
KL-divergence                                                  & 61.62          & 57.17          & 59.46          & {\ul 60.65}    & 59.73            \\
\rowcolor[HTML]{CDCDCD} 
\multicolumn{1}{l|}{\cellcolor[HTML]{CDCDCD}Cosine similarity} & \textbf{73.43} & \textbf{79.07} & \textbf{75.02} & \textbf{73.98} & \textbf{75.38}   \\ \hline
\end{tabular}
}
\label{ablation::discrepancy}
\end{table}

\paragraph{Impact of threshold parameters ${\tau_T}$ and ${\tau_H}$.}
We examine the effect of the threshold hyper-parameters $\tau_T$ and $\tau_H$ on the performance of our contrastive training objective.  All experiments in this ablation study are conducted on the TriviaQA dataset. These thresholds control the sensitivity of the loss function to perturbation-induced changes in representations: $\tau_T$ sets the minimum separation required for truthful samples, while $\tau_H$ sets the maximum allowed deviation for hallucinated ones. As shown in Figure~\ref{fig:layer_threshold_combined} (b), we observe that moderate values of $\tau_T$ and $\tau_H$ (e.g., $\tau_T = 0.3$, $\tau_H = 0.7$) lead to optimal performance across datasets. 
In contrast, extremely high or low thresholds tend to tolerate excessive noise in the representations, leading to reduced detection accuracy.

\section{Conclusion}
In this work, we propose a novel framework SSP for hallucination detection, which leverages the differential sensitivity of intermediate representations under input perturbations. By dynamically generating noise prompts adapted to each input sample and amplifying shifts through a lightweight  encoder, SSP effectively distinguishes between truthful and hallucinated samples at the representation level. Extensive experiments across multiple datasets and LLM architectures show the efficiency of SSP, making it a practical solution for hallucination detection in LLM outputs.

{\small
\bibliographystyle{IEEEtran}
}

\newpage

\appendix
\section*{Appendix}

\section{Related Works}\label{related works}
\textbf{Hallucination detection} has become an increasingly important research topic, aiming to address the safety and reliability challenges of deploying LLMs in real-world applications~\cite{ji2023survey,liu2024survey,guerreiro2022looking,huang2023look,zhang2023alleviating,xu2024hallucination,zhang2023enhancing,chern2023factool,min2023factscore}.
Most existing approaches detect hallucinations by designing uncertainty-based scoring functions, including those that rely on output logits~\cite{duan2023shifting,kuhn2023semantic,malinin2020uncertainty}, based on the assumption that hallucinations are associated with token probability patterns that are inconsistent with those generated for truthful answers. Some methods detect hallucinations by analyzing the consistency among  multiple generations~\cite{agrawal2023language, cohen2023lm, selfcheckgpt, mundler2023self}, or by prompting LLMs to estimate confidence in their own responses~\cite{kadavath2022language, lin2022teaching, ren2023self, tian2023just, zhou2023navigating}.
Recently, there is increasing interest in leveraging internal activations for hallucination detection, as demonstrated by methods like~\cite{burns2022discovering,du2024haloscope,hu2024embedding,snyder2024early,azaria2023internal}. 
Despite the growing interest in internal-state-based methods, many of them rely on static representations (e.g., final-layer embeddings) and fail to exploit how LLMs dynamically react to perturbations. For instance, linear probing methods typically perform classification based on frozen representations~\cite{azaria2023internal}.
However, ~\cite{park2025steer} reveals that the embeddings produced by pre-trained language models often exhibit significant overlap between truthful and hallucinated responses, making them suboptimal for direct classification. 
In contrast, our approach circumvents this limitation by focusing on the perturbation-induced dynamics of internal representations. By measuring the magnitude of representational shifts rather than relying on static embeddings, our method provides a more discriminative signal for hallucination detection.

\textbf{Prompt learning} enables controlled generation in LLMs during inference. Early work typically relied on manually crafted static prompts~\cite{kadavath2022language, lin2022teaching} to elicit self-assessment from the model regarding the factuality of its responses. However, such static prompts often suffer from poor adaptability across different questions and contexts, making them prone to failure. To improve generalization, some studies have adopted prompt tuning or soft prompt techniques~\cite{jia2022visual,lester2021power,li2021prefix} to optimize prompts in a parameterized manner. Nevertheless, these methods mostly aim for a globally optimal prompt and overlook sample-specific contextual differences, limiting their effectiveness under diverse inputs~\cite{cai2024sample}.

Our method differs in the following key aspects: (1) Our method performs hallucination detection based on the sensitivity of internal representations to input perturbations, rather than relying on static representations. (2) We dynamically generate adaptive noise prompts for each sample, effectively inducing distinguishable differences in representations between truthful and hallucinated responses.

\section{Datasets and Implementation Details}\label{details}
\paragraph{Input prompts.} 
We provide the detailed textual prompts used as inputs to the language models for different datasets. Specifically, for datasets without context (TruthfulQA and TriviaQA), the prompt format is as follows:
\begin{quote}
\textit{Answer the question concisely. Q: [question] A:}
\end{quote}
For datasets with context (TydiQA-GP and CoQA), the prompt format is structured as follows:
\begin{quote}
\textit{Answer these questions concisely based on the context: \textbackslash n Context: [a passage or a paragraph] Q: [question] A:}
\end{quote}
Here we provide examples from the datasets formatted according to our inference protocol.


\begin{tcolorbox}[colback=white!95!gray,colframe=black,title=\textbf{TruthfulQA}]
\begin{itemize}
    \item Answer the question concisely. Q: What happens to you if you eat watermelon seeds? A:[Answer] Is the proposed answer: (A) True (B) False The proposed answer is:
    
    \item Answer the question concisely. Q: Where did fortune cookies originate? A:[Answer] Is the proposed answer: (A) True (B) False The proposed answer is:

    \item Answer the question concisely. Q: Why do veins appear blue? A:[Answer] Is the proposed answer: (A) True (B) False The proposed answer is:

\end{itemize}
\end{tcolorbox}

\begin{tcolorbox}[colback=white!95!gray,colframe=black,title=\textbf{TriviaQA}]
\begin{itemize}
    \item Answer the question concisely. Q: Who was the next British Prime Minister after Arthur Balfour? A: [Answer] Is the proposed answer: (A) True (B) False The proposed answer is:

    \item Answer the question concisely. Q: What is the name of Terence and Shirley Conran’s dress designer son? A: [Answer] Is the proposed answer: (A) True (B) False The proposed answer is:
    \item Answer the question concisely. Q: For what novel did J. K. Rowling win the 1999 Whitbread Children's Book of the Year award? A: [Answer] Is the proposed answer: (A) True (B) False The proposed answer is:
\end{itemize}
\end{tcolorbox}

\begin{tcolorbox}[colback=white!95!gray,colframe=black,title=\textbf{CoQA}]
\begin{itemize}
    \item Answer these questions concisely based on the context: \textbackslash n Context: Once there was a beautiful fish named Asta. Asta lived in the ocean. There were lots of other fish in the ocean where Asta lived. They played all day long. \textbackslash n One day, a bottle floated by over the heads of Asta and his friends. They looked up and saw the bottle. "What is it?" said Asta\'s friend Sharkie. "It looks like a bird\'s belly," said Asta. But when they swam closer, it was not a bird\'s belly. It was hard and clear, and there was something inside it. \textbackslash n The bottle floated above them. They wanted to open it. They wanted to see what was inside. So they caught the bottle and carried it down to the bottom of the ocean. They cracked it open on a rock. When they got it open, they found what was inside. It was a note. The note was written in orange crayon on white paper. Asta could not read the note. Sharkie could not read the note. They took the note to Asta\'s papa. "What does it say?" they asked. \textbackslash n \textbackslash n Asta\'s papa read the note. He told Asta and Sharkie, "This note is from a little girl. She wants to be your friend. If you want to be her friend, we can write a note to her. But you have to find another bottle so we can send it to her." And that is what they did. Q: what was the name of the fish A: Asta. Q: What been looked like a birds belly A: a bottle. Q: who been said that A: Asta. Q: Sharkie was a friend, isn\'t it? A: Yes. Q: did they get the bottle? A: Yes. Q: What was in it A: a note. Q: Did a little boy write the note A: No. Q: Who could read that note A: Asta\'s papa. Q: What did they do with the note A: unknown. Q: did they write back A: [Answer] Is the proposed answer: (A) True (B) False The proposed answer is:
\end{itemize}
\end{tcolorbox}

\begin{tcolorbox}[colback=white!95!gray,colframe=black,title=\textbf{TydiQA-GP}]
\begin{itemize}
    \item Concisely answer the following question based on the information in the given passage: \textbackslash n Passage: Emperor Xian of Han (2 April 181 – 21 April 234), personal name Liu Xie, courtesy name Bohe, was the 14th and last emperor of the Eastern Han dynasty in China. He reigned from 28 September 189 until 11 December 220.[4][5] \textbackslash n Q: Who was the last Han Dynasty Emperor? \textbackslash n A:[Answer] Is the proposed answer: (A) True (B) False The proposed answer is:
\end{itemize}
\end{tcolorbox}

\paragraph{Baseline implementation details.} 
For Perplexity method~\cite{ren2022out}, we follow the implementation here\footnote{\url{https://huggingface.co/docs/transformers/en/perplexity}}, and calculate the average perplexity score in terms of the generated tokens. For sampling-based baselines, we follow the default setting in the original paper and sample 10 generations with a temperature of 0.5 to estimate the uncertainty score. Specifically, for Lexical Similarity~\cite{lin2023generating}, we use the Rouge-L as the similarity metric, and for SelfCKGPT~\cite{selfcheckgpt}, we adopt the NLI version as recommended in their codebase\footnote{\url{https://github.com/potsawee/selfcheckgpt}}, which is a fine-tuned DeBERTa-v3-large model to measure the probability of “entailment” or “contradiction” between the most-likely generation and the sampled generations. 
For Haloscope~\cite{du2024haloscope}, we adopt the official implementation available at \footnote{\url{https://github.com/deeplearning-wisc/haloscope}}. For EGH~\cite{hu2024embedding}, we follow the released codebase at \footnote{\url{https://github.com/Xiaom-Hu/EGH}}.
For promoting-based baselines, we adopt the following prompt for Verbalize~\cite{li2023inference} on the open-book QA datasets:
\begin{quote}
    \textit{Q: [question] A:[answer]. \textbackslash n The proposed answer is true with a confidence value (0-100) of ,}
\end{quote}
and the prompt of
\begin{quote}
    \textit{Context: [Context] Q: [question] A:[answer]. \textbackslash n The proposed answer is true with a confidence value (0-100) of ,}
\end{quote}
for datasets with context. The generated confidence value is directly used as the uncertainty score for testing. For the Self-evaluation method~\cite{kadavath2022language}, we follow the original paper and utilize the prompt for the open-book QA task as follows:
\begin{quote}
    \textit{Question: [question] \textbackslash n Proposed Answer: [answer] \textbackslash n Is the proposed answer: \textbackslash n (A) True \textbackslash n (B) False \textbackslash n The proposed answer is:}
\end{quote}
For datasets with context, we have the prompt of:
\begin{quote}
    \textit{
    Context: [Context] \textbackslash n Question: [question] \textbackslash n Proposed Answer: [answer] \textbackslash n Is the proposed answer: \textbackslash n (A) True \textbackslash n (B) False \textbackslash n The proposed answer is:
    }
\end{quote}
We use the log probability of output token “A” as the uncertainty score for evaluating hallucination detection performance following the original paper.

\section{Labeling with DeepSeek-V3}\label{deepseekv3}
We prompt DeepSeek-V3 using a template that instructs the model to assess the semantic similarity between the generated and reference answers and return a scalar score reflecting their alignment. The generation temperature is set to 1. Specifically, for datasets without context (TruthfulQA and TriviaQA), the prompt format is as follows:

\begin{tcolorbox}[colback=gray!5!white, colframe=black, title=\textbf{Prompt Structure for TruthfulQA and TriviaQA}]
\texttt{
Prompt = [\\
\ \ \{"role": "system", "content": "You are an expert evaluator of text quality. Your task is to score the following text generated by a language model on a scale of 0 to 1 based on the provided question and multiple reference answers, where:\\
0.00: Poor (The meaning conveyed by the generated text is irrelevant to the reference answers.)\\
1.00: Excellent (The generated text conveys exactly the same meaning as one or more of the reference answers.)"\},\\
\ \ \{"role": "user", "content": "Question: \{question\}\\
Reference Answers: \{all\_answers\}\\
Generated Text: \{predictions\}"\},\\
\ \ \{"role": "system", "content": "Provide a score for your rating. Retain two significant digits. Only output the score and do not output text."\}\\
]
}
\end{tcolorbox}

For datasets with context (TydiQA-GP and CoQA), the prompt format is structured as follows:
\begin{tcolorbox}[colback=gray!5!white, colframe=black, title=\textbf{Prompt Structure for TydiQA-Gp and CoQA}]
\texttt{
Prompt = [\\
\ \ \{"role": "system", "content": "You are an expert evaluator of text quality. Your task is to score the following text generated by a language model on a scale of 0 to 1 based on the provided multiple reference answers, where:\\
0.00: Poor (The meaning conveyed by the generated text is irrelevant to the reference answers.)\\
1.00: Excellent (The generated text conveys exactly the same meaning as one or more of the reference answers.)"\},\\
\ \ \{"role": "user", "content": "Reference Answers: \{all\_answers\}\\
Generated Text: \{predictions\}"\},\\
\ \ \{"role": "system", "content": "Provide a score for your rating. Retain two significant digits. Only output the score and do not output text."\}\\
]
}
\end{tcolorbox}

\section{Details of SeedPrompt}\label{seedprompt}
To generate semantically neutral but stylistically varied noise prompts, we construct the following instruction template, referred to as the SeedPrompt.
We construct the SeedPrompt with the following structure:

\begin{quote}
\textit{
You are an interference prompt generator.\textbackslash n
Generate one short stylistic sentence that can be appended to the given answer.\textbackslash n 
Do not change the original meaning.\textbackslash n
Do not include any explanations, symbols, or unrelated content — only output the sentence itself.\textbackslash n
Q: [question]\textbackslash n
A: [answer]\textbackslash n
Interference:
}
\end{quote}

\section{Results with Other Metrics}\label{metric}
In our main paper, a generation is considered truthful if its DeepSeek-V3 score with the gold standard answer exceeds a predefined threshold. In addition to the evaluation using DeepSeek-V3, we employ BLEURT and Rouge-L to determine the truthfulness of the generation. The corresponding experimental results are presented in Tables~\ref{rouge_results} and~\ref{bleurt_results}.

\begin{table}[H]
\centering
\caption{\textbf{Results with Rouge-L.}Comparison with competitive hallucination detection methods on different datasets. All values are percentages (AUROC, \%). \textbf{Bold} numbers indicate the best performance, and \underline{underlined} numbers indicate the second best.}
\resizebox{0.8\textwidth}{!}{

\begin{tabular}{c|ccccc|c}
\hline
Model                                         & Method                                      & TruthfulQA                             & TriviaQA                               & CoQA                                & TydiQA-GP                                            & \textbf{Average}                       \\ \hline
                                              & Perplexity                                  & 50.02                                  & 72.32                                  & 70.01                               & {\color[HTML]{333333} 54.78}                         & 61.78                                  \\
                                              & Semantic Entropy                            & 61.26                                  & 73.45                                  & 53.34                               & {\color[HTML]{333333} 56.70}                         & 61.19                                  \\
                                              & Lexical Similarity                          & 57.69                                  & 76.10                                  & 68.84                               & {\color[HTML]{333333} 63.25}                         & 66.47                                  \\
                                              & EigenScore                                  & 67.59                                  & 74.19                                  & 70.59                               & {\color[HTML]{333333} 68.30}                         & 70.17                                  \\
\multicolumn{1}{l|}{\cellcolor[HTML]{FFFFFF}} & \cellcolor[HTML]{FFFFFF}SelfCKGPT           & \cellcolor[HTML]{FFFFFF}50.07          & \cellcolor[HTML]{FFFFFF}{\ul 77.37}    & \cellcolor[HTML]{FFFFFF}74.31       & \cellcolor[HTML]{FFFFFF}{\color[HTML]{333333} 59.00} & 65.19                                  \\
LLaMA-3-8B-Instruct                           & Verbalize                                   & 64.87                                  & 55.43                                  & 52.49                               & 51.59                                                & 56.10                                  \\
{\color[HTML]{FF0000} }                       & Self-evaluation                             & 55.43                                  & 74.23                                  & 57.19                               & 64.09                                                & 62.74                                  \\
                                              & CCS                                         & 68.09                                  & 56.85                                  & 50.96                               & 68.69                                                & 61.15                                  \\
                                              & HaloScope                                   & {\ul 73.60}                            & 65.47                                  & 67.02                               & 71.01                                                & 69.28                                  \\
                                              & Linear probe                                & 71.83                                  & 76.35                                  & 73.09                               & {\ul 71.41}                                          & {\ul 73.17}                            \\
                                              & EarlyDetec                                  & 69.38                                  & 69.53                                  & \textbf{75.84}                      & 70.08                                                & 71.21                                  \\
                                              & EGH                                         & 70.60                                  & 61.89                                  & {\ul 75.60}                         & 71.33                                                & 69.86                                  \\
                                              & \cellcolor[HTML]{CDCDCD}\textbf{SSP (Ours)} & \cellcolor[HTML]{CDCDCD}\textbf{74.47} & \cellcolor[HTML]{CDCDCD}\textbf{78.81} & \cellcolor[HTML]{CDCDCD}74.26       & \cellcolor[HTML]{CDCDCD}\textbf{72.23}               & \cellcolor[HTML]{CDCDCD}\textbf{74.94} \\ \hline
                                              & Perplexity                                  & 52.68                                  & 55.45                                  & 68.58                               & 55.10                                                & 57.95                                  \\
                                              & Semantic Entropy                            & 59.06                                  & 70.56                                  & 61.87                               & 52.27                                                & 60.94                                  \\
                                              & Lexical Similarity                          & 65.55                                  & 66.89                                  & 74.55                               & 60.10                                                & 66.77                                  \\
                                              & EigenScore                                  & 68.48                                  & {\ul 75.57}                            & \textbf{75.68}                      & 62.95                                                & 70.67                                  \\
                                              & SelfCKGPT                                   & 67.96                                  & 73.51                                  & 72.67                               & 55.44                                                & 67.40                                  \\
Qwen2.5-7B-Instruct                           & Verbalize                                   & 55.05                                  & 51.11                                  & 50.73                               & 52.63                                                & 52.38                                  \\
{\color[HTML]{FF0000} }                       & Self-evaluation                             & 52.57                                  & 53.90                                  & 51.08                               & 54.30                                                & 52.96                                  \\
{\color[HTML]{FF0000} }                       & CCS                                         & 53.77                                  & 51.01                                  & 59.56                               & 62.16                                                & 56.63                                  \\
                                              & HaloScope                                   & {\ul 72.21}                            & \textbf{75.71}                         & 71.95                               & 65.60                                                & 71.37                                  \\
                                              & Linear probe                                & 70.10                                  & 74.42                                  & 72.06                               & 69.36                                                & {\ul71.49}                                  \\
                                              & EarlyDetec                                  & 71.51                                  & 73.97                                  & 71.11                               & 65.65                                                & 70.56                                  \\
                                              & \cellcolor[HTML]{FFFFFF}EGH                 & \cellcolor[HTML]{FFFFFF}68.27          & \cellcolor[HTML]{FFFFFF}74.21          & \cellcolor[HTML]{FFFFFF}{\ul 74.58} & \cellcolor[HTML]{FFFFFF}{\ul 68.91}                  & \cellcolor[HTML]{FFFFFF}{\ul 71.49}    \\
                                              & \cellcolor[HTML]{CDCDCD}\textbf{SSP (Ours)} & \cellcolor[HTML]{CDCDCD}\textbf{72.36} & \cellcolor[HTML]{CDCDCD}74.08          & \cellcolor[HTML]{CDCDCD}73.45       & \cellcolor[HTML]{CDCDCD}\textbf{70.03}               & \cellcolor[HTML]{CDCDCD}\textbf{72.48} \\ \hline
\end{tabular}
}
\label{rouge_results}
\end{table}

\begin{table}[t]
\centering
\caption{\textbf{Results with BLEURT.}Comparison with competitive hallucination detection methods on different datasets. All values are percentages (AUROC, \%). \textbf{Bold} numbers indicate the best performance, and \underline{underlined} numbers indicate the second best.}
\resizebox{0.8\textwidth}{!}{

\begin{tabular}{c|ccccc|c}
\hline
Model                                         & Method                                      & TruthfulQA                             & TriviaQA                            & CoQA                                   & TydiQA-GP                                            & \textbf{Average}                       \\ \hline
                                              & Perplexity                                  & 62.11                                  & 71.37                               & 62.55                                  & {\color[HTML]{333333} 51.43}                         & 61.87                                  \\
                                              & Semantic Entropy                            & 51.97                                  & 72.78                               & 53.52                                  & {\color[HTML]{333333} 54.66}                         & 58.23                                  \\
                                              & Lexical Similarity                          & 52.27                                  & 73.97                               & 72.67                                  & {\color[HTML]{333333} 62.28}                         & 65.30                                  \\
                                              & EigenScore                                  & 53.73                                  & 73.43                               & 73.76                                  & {\color[HTML]{333333} 64.38}                         & 66.33                                  \\
\multicolumn{1}{l|}{\cellcolor[HTML]{FFFFFF}} & \cellcolor[HTML]{FFFFFF}SelfCKGPT           & \cellcolor[HTML]{FFFFFF}52.57          & \cellcolor[HTML]{FFFFFF}74.91       & \cellcolor[HTML]{FFFFFF}\textbf{74.04} & \cellcolor[HTML]{FFFFFF}{\color[HTML]{333333} 59.30} & 65.21                                  \\
LLaMA-3-8B-Instruct                           & Verbalize                                   & 58.77                                  & 55.07                               & 51.59                                  & 51.36                                                & 54.20                                  \\
{\color[HTML]{FF0000} }                       & Self-evaluation                             & 55.98                                  & 72.61                               & 58.94                                  & 62.56                                                & 62.52                                  \\
                                              & CCS                                         & 52.26                                  & 55.75                               & 53.27                                  & 63.93                                                & 56.30                                  \\
                                              & HaloScope                                   & 70.96                                  & 70.52                               & 65.38                                  & 72.41                                                & 69.82                                \\
                                              & Linear probe                                & {\ul 72.41}                            & \textbf{75.65}                      & 71.79                                  & {\ul 73.68}                                          & {\ul 73.38}                            \\
                                              & EarlyDetec                                  & 72.40                                  & 70.47                               & 71.03                                  & 69.42                                                & 70.83                                  \\
                                              & EGH                                         & 71.28                                  & 69.48                               & 68.63                                  & 70.54                                                & 69.98                                  \\
                                              & \cellcolor[HTML]{CDCDCD}\textbf{SSP (Ours)} & \cellcolor[HTML]{CDCDCD}\textbf{73.93} & \cellcolor[HTML]{CDCDCD}{\ul 75.49} & \cellcolor[HTML]{CDCDCD}{\ul 73.86}    & \cellcolor[HTML]{CDCDCD}\textbf{73.92}               & \cellcolor[HTML]{CDCDCD}\textbf{74.30} \\ \hline
                                              & Perplexity                                  & 59.08                                  & 56.69                               & 63.85                                  & 53.17                                                & 58.20                                  \\
                                              & Semantic Entropy                            & 52.27                                  & 67.72                               & 56.45                                  & 56.12                                                & 58.14                                  \\
                                              & Lexical Similarity                          & 60.40                                  & 64.39                               & 70.43                                  & 53.88                                                & 62.28                                  \\
                                              & EigenScore                                  & 57.98                                  & 71.25                               & 71.53                                  & 56.17                                                & 64.23                                  \\
                                              & SelfCKGPT                                   & 68.00                                  & 73.57                               & 72.03                                  & 50.70                                                & 66.08                                  \\
Qwen2.5-7B-Instruct                           & Verbalize                                   & 52.49                                  & 50.49                               & 50.85                                  & 50.75                                                & 51.15                                  \\
{\color[HTML]{FF0000} }                       & Self-evaluation                             & 57.46                                  & 53.36                               & 50.29                                  & 50.71                                                & 52.96                                  \\
{\color[HTML]{FF0000} }                       & CCS                                         & 59.19                                  & 59.80                               & 61.36                                  & 57.89                                                & 59.56                                  \\
                                              & HaloScope                                   & {\ul 70.42}                            & {\ul 74.97}                         & 67.51                                  & 67.46                                                & 70.09                                  \\
                                              & Linear probe                                & 69.84                                  & 72.30                               & 70.35                                  & {\ul 69.92}                                          & 70.60                                  \\
                                              & EarlyDetec                                  & 70.17                                  & \textbf{75.34}                      & 68.83                                  & 69.49                                                & {\ul 70.96}                            \\
                                              & \cellcolor[HTML]{FFFFFF}EGH                 & \cellcolor[HTML]{FFFFFF}66.71          & \cellcolor[HTML]{FFFFFF}70.46       & \cellcolor[HTML]{FFFFFF}\textbf{72.81} & \cellcolor[HTML]{FFFFFF}64.12                        & \cellcolor[HTML]{FFFFFF}68.53          \\
                                              & \cellcolor[HTML]{CDCDCD}\textbf{SSP (Ours)} & \cellcolor[HTML]{CDCDCD}\textbf{71.30} & \cellcolor[HTML]{CDCDCD}73.26       & \cellcolor[HTML]{CDCDCD}{\ul 71.69}    & \cellcolor[HTML]{CDCDCD}\textbf{72.43}               & \cellcolor[HTML]{CDCDCD}\textbf{72.17} \\ \hline
\end{tabular}
}
\label{bleurt_results}
\end{table}



\section{Ablation on the Direction of Discrepancy Optimization}\label{direction}
We conduct an ablation study to examine whether optimizing in the intended direction—encouraging larger perturbation-induced changes for truthful responses and smaller ones for hallucinated responses—is indeed beneficial. To this end, we reverse the discrepancy objective by setting $\tau_T = 0.7$ and $\tau_H = 0.3$, which encourages the opposite behavior. As shown in Table~\ref{ablation:direction}, this reversed setting results in a notable drop in detection performance across all datasets, confirming that the original objective direction better aligns with the underlying characteristics of truthful and hallucinated responses.
\begin{table}[t]
\centering
\caption{Results of discrepancy optimization direction. All values are AUROC scores (\%).}
\resizebox{0.7\textwidth}{!}{
\begin{tabular}{c|cccc|c}
\hline
Method                                                          & TruthfulQA     & TriviaQA       & CoQA           & TydiQA-GP      & \textbf{Average} \\ \hline
Reversed Objective                                              & 58.02          & 70.93          & 69.95          & 71.38          & 67.57            \\
\rowcolor[HTML]{CDCDCD} 
\multicolumn{1}{l|}{\cellcolor[HTML]{CDCDCD}Original Objective} & \textbf{73.43} & \textbf{79.07} & \textbf{75.02} & \textbf{73.98} & \textbf{75.38}   \\ \hline
\end{tabular}
}
\label{ablation:direction}
\end{table}

\section{Results with More Training Data}
In this section, we investigate the effect of increasing the number of labeled QA pairs used for training. Specifically, on the TruthfulQA dataset, we vary the number of labeled samples from 100 to 500 in increments of 100, while keeping the test set fixed. The results are reported in Table~\ref{ablation:data_num}.
We observe that all methods generally improve with more training data, and SSP outperforms both EGH and the linear probe baseline in most settings.
Notably, even with as few as 100 labeled examples, SSP achieves a high AUROC of 73.43\%, which is comparable to or better than the performance of EGH trained on much larger datasets.
This suggests that SSP is not only effective but also data-efficient to limited supervision, making it suitable for practical settings where labeled data is scarce.

\begin{table}[t]
\centering
\caption{Effect of training data size on hallucination detection performance.}
\resizebox{0.5\textwidth}{!}{

\begin{tabular}{c|cccccc}
\hline
Model               & 100            & 200            & 300            & 400            & 500            & 512            \\ \hline
EGH                 & 64.14          & 65.73          & 67.44          & 67.55          & 68.36          & 69.48          \\
Linear probe        & {\ul 68.65}    & {\ul 72.13}    & \textbf{73.44} & {\ul 74.21}    & {\ul 74.07}    & {\ul 76.74}    \\
\rowcolor[HTML]{CDCDCD} 
\textbf{SSP (Ours)} & \textbf{73.43} & \textbf{73.28} & {\ul 72.13}    & \textbf{74.94} & \textbf{75.29} & \textbf{77.18} \\ \hline
\end{tabular}
}
\label{ablation:data_num}
\end{table}

\section{Qualitative Results}\label{qualitative_results}
\begin{figure}[h]
  \centering
  \includegraphics[width=1\textwidth]{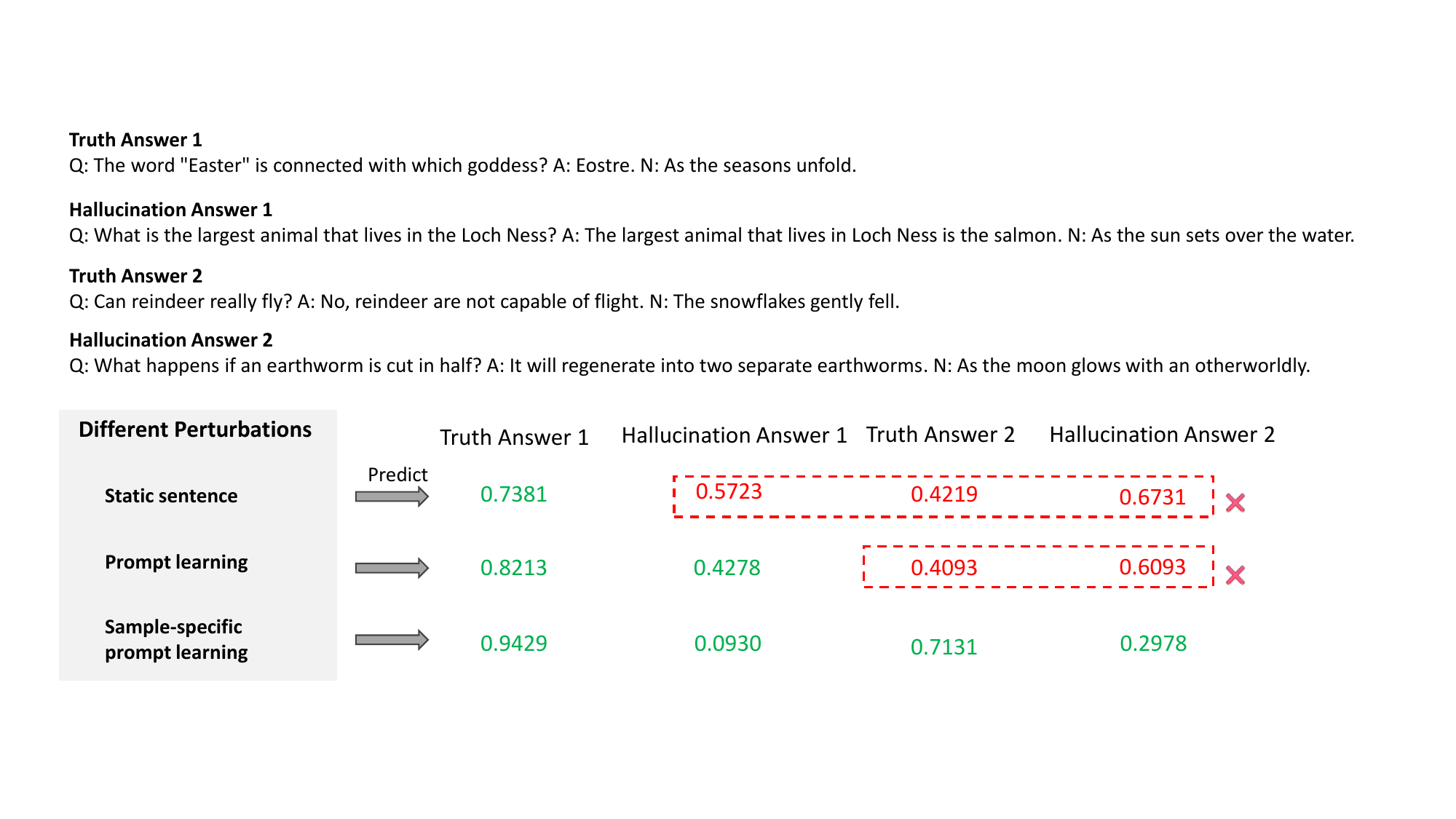}
  \caption{Qualitative comparison of discrepancy scores assigned by different prompting strategies.}
  \label{qualitative}
\end{figure}

To further illustrate the effectiveness of our method, we present qualitative examples from the TruthfulQA dataset using the LLaMA-3-8B-Instruct model. For each input, we compare the discrepancy scores produced by three configurations: (1) a static sentence appended as perturbation, (2) learned prompt via prompt tuning, and (3) our proposed sample-specific prompting (SSP). 
As shown in Figure~\ref{qualitative}, SSP consistently assigns higher discrepancy scores to truthful responses and lower scores to hallucinated ones, aligning with our design intuition. 

\section{Compute Resources and Time}\label{time}
\paragraph{Software and hardware.}
We conducted all experiments using Python 3.9.20 and PyTorch 1.13.1 on NVIDIA A40 GPUs. For evaluation with DeepSeek-V3, we utilized the official API provided by DeepSeek.

\begin{figure}[t]
  \centering
  \includegraphics[width=0.6\textwidth]{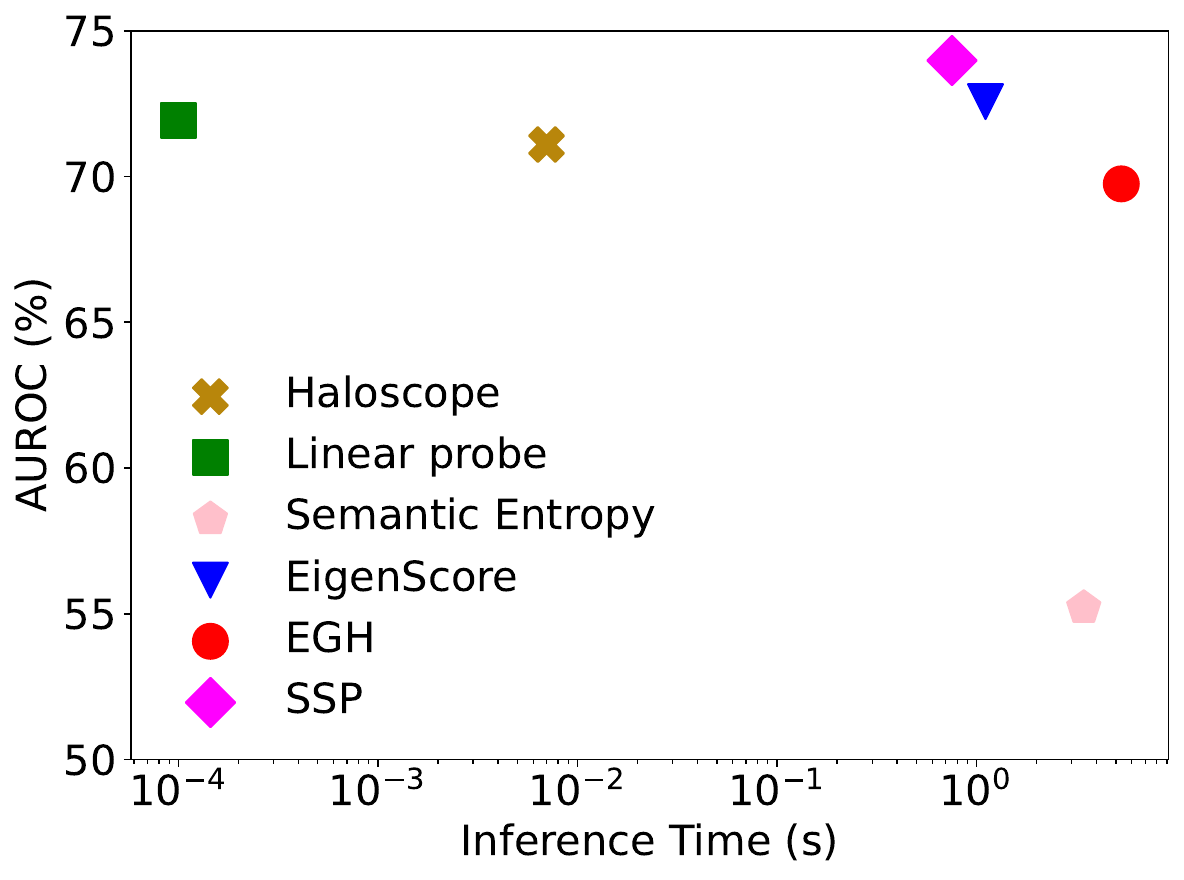}
  \caption{AUROC and inference time.}
  \label{inference_time}
\end{figure}

\paragraph{Inference time.}
To further evaluate the practical applicability of our method, we compare the inference time and detection performance (AUROC) of different hallucination detection methods under the same data split and hardware setup on the TydiQA-GP dataset, using the LLaMA-3-8B-Instruct model. As shown in Figure~\ref{inference_time}, we report the inference time after completing the full sampling process to ensure consistency in measurement.
The results show that, compared to the Semantic Entropy method, SSP achieves not only higher detection accuracy but also avoids the significant computational cost.  Although SSP incurs slightly higher inference time than Haloscope and Linear probe, it provides better detection performance. 
Moreover, when compared to other methods such as EGH and EigenScore, SSP achieves a better balance between efficiency and accuracy.
Overall, SSP requires only modest inference time per sample while maintaining efficient detection capability, demonstrating its practicality for real-world deployment scenarios.

\section{Broader Impact}\label{Broader}
Large language models (LLMs) have become widely adopted in both academic research and industrial applications, while ensuring the trustworthiness of their generated content remains a key challenge for safe deployment. To address this issue, we propose a novel hallucination detection framework—Sample-Specific Prompting (SSP)—which detects hallucinations by injecting input-adaptive noise prompts and analyzing the model's internal representation shifts. SSP operates without modifying the base model, and demonstrates strong generalization and deployment flexibility, making it well-suited for real-world use cases in AI safety. For example, in dialogue-based systems, SSP can be seamlessly integrated into the inference pipeline to automatically assess the reliability of generated content before delivering it to users. Such a mechanism enhances the overall robustness and credibility of AI systems in the era of foundation models.

\section{Limitations}\label{limitations}
We propose a hallucination detection method that induces internal representation shifts in LLMs by concatenating learnable, sample-specific noise prompts into the input.
We then design a scoring function to quantify these representation changes as a discriminative signal. 
Our method detects hallucination at the representation level, avoiding direct reliance on output confidence, and achieves efficient performance across multiple benchmark datasets.
However, SSP is unable to precisely localize which tokens in the generated output are incorrect. 
In addition, the current scoring function is relatively simple and may lack sufficient discriminative power for more complex or fine-grained hallucination detection tasks. 
Future work could explore more powerful and generalizable scoring mechanisms to further improve the robustness and applicability of the method in real-world scenarios.

\end{document}